%% file: main.tex
\def\year{2022}\relax
\documentclass[letterpaper]{article} 
\usepackage{aaai22}  
\usepackage{times}  
\usepackage{helvet}  
\usepackage{courier}  
\usepackage[hyphens]{url}  
\usepackage{graphicx} 
\usepackage{natbib}  
\usepackage{caption} 
\DeclareCaptionStyle{ruled}{labelfont=normalfont,labelsep=colon,strut=off} 
\usepackage[title]{appendix}

\usepackage{algorithm, multirow}
\usepackage{algorithmic}
\usepackage[table,xcdraw]{xcolor}
\usepackage{subfig,amsmath, amssymb}
\usepackage{wrapfig}
\newcommand{\name}{$\Delta$-UQ~}
\usepackage{multirow}

\usepackage{newfloat}
\usepackage{listings}
\floatstyle{ruled}
\newfloat{listing}{tb}{lst}{}
\floatname{listing}{Listing}
\nocopyright
\title{$\Delta$-UQ: Accurate Uncertainty Quantification via Anchor Marginalization}
\author{Rushil Anirudh \thanks{equal contribution} and Jayaraman J. Thiagarajan\footnotemark[1]}
\affiliations{{\tt \{anirudh1, jjayaram\}@llnl.gov}\\
Center for Applied Scientific Computing (CASC)\\
Lawrence Livermore National Laboratory.}
\begin{document}

\maketitle

\begin{abstract}
\emph{We present \name -- a novel, general-purpose uncertainty estimator using the concept of \emph{anchoring} in predictive models. Anchoring works by first transforming the input into a tuple consisting of an anchor point drawn from a prior distribution, and a combination of the input sample with the anchor using  a pretext encoding scheme. This encoding is such that the original input can be perfectly recovered from the tuple -- regardless of the choice of the anchor. Therefore, any predictive model should be able to predict the target response from the tuple alone (since it implicitly represents the input). Moreover, by varying the anchors for a fixed sample, we can estimate uncertainty in the prediction even using only a single predictive model. We find this uncertainty is deeply connected to improper sampling of the input data, and inherent noise, enabling us to estimate the total uncertainty in any system. With extensive empirical studies on a variety of use-cases, we demonstrate that \name outperforms several competitive baselines. Specifically, we study model fitting, sequential model optimization, model based inversion in the regression setting and out of distribution detection, \& calibration under distribution shifts for classification.\let\thefootnote\relax \footnote{This work was performed under the auspices of the U.S. Department of Energy by Lawrence Livermore National Laboratory under Contract DE-AC52-07NA27344}}
\end{abstract}

\section{Introduction}
\input{intro.tex}

\section{Proposed Approach}
\input{methods.tex}
\section{Related Work}
\input{related.tex}

\section{Experiments}
\input{expts.tex}
\section{Conclusion and Discussion}
In this paper, we presented a simple new general-purpose, single model uncertainty estimator that does not need require any additional training modules or specific regularization strategies. We introduced the concept of anchoring-based uncertainty estimation,  that re-parameterize an input sample into a tuple consisting of the anchor, which is a random sample drawn from a prior distribution, and a pretext encoding of the input. By marginalizing out the effect of the choice of different anchors on the predictions, we are able to accurately estimate both aleatoric and epistemic uncertainties in the system.  We show that our method, referred to as \name, produces meaningful uncertainties for both regression and classification tasks and are consistently superior to several existing uncertainty estimators.

The idea of anchoring opens several interesting future directions of work. For simplicity during training, we fix the number of anchors to be 1, which computes a loss only on point estimates. However, it is conceivable that by cleverly choosing a larger number of anchors at train time, the resulting mean prediction itself can be made more robust. Moreover, the effect of the choice of the anchor distribution, $P(\mathcal{R})$ might play a key role in determining both prediction fidelity and uncertainty calibration. Here we fixed $P(\mathcal{R}) = P(\mathcal{X})$, but better anchor priors can be designed which may further improve performance. Finally, anchoring also presents interesting avenues for theoretical analysis to understand its connection to reconstruction-based representation learning approaches such as~\cite{falcon2021aavae} and~\cite{sinha2021consistency}.

\bibliography{refs}
\newpage
\begin{appendices}
\section{Illustrations}
\begin{figure*}[!htb]
\centering
\subfloat[][An overview of \name]{\includegraphics[width=0.4\linewidth,clip,trim=0 0 0 0]{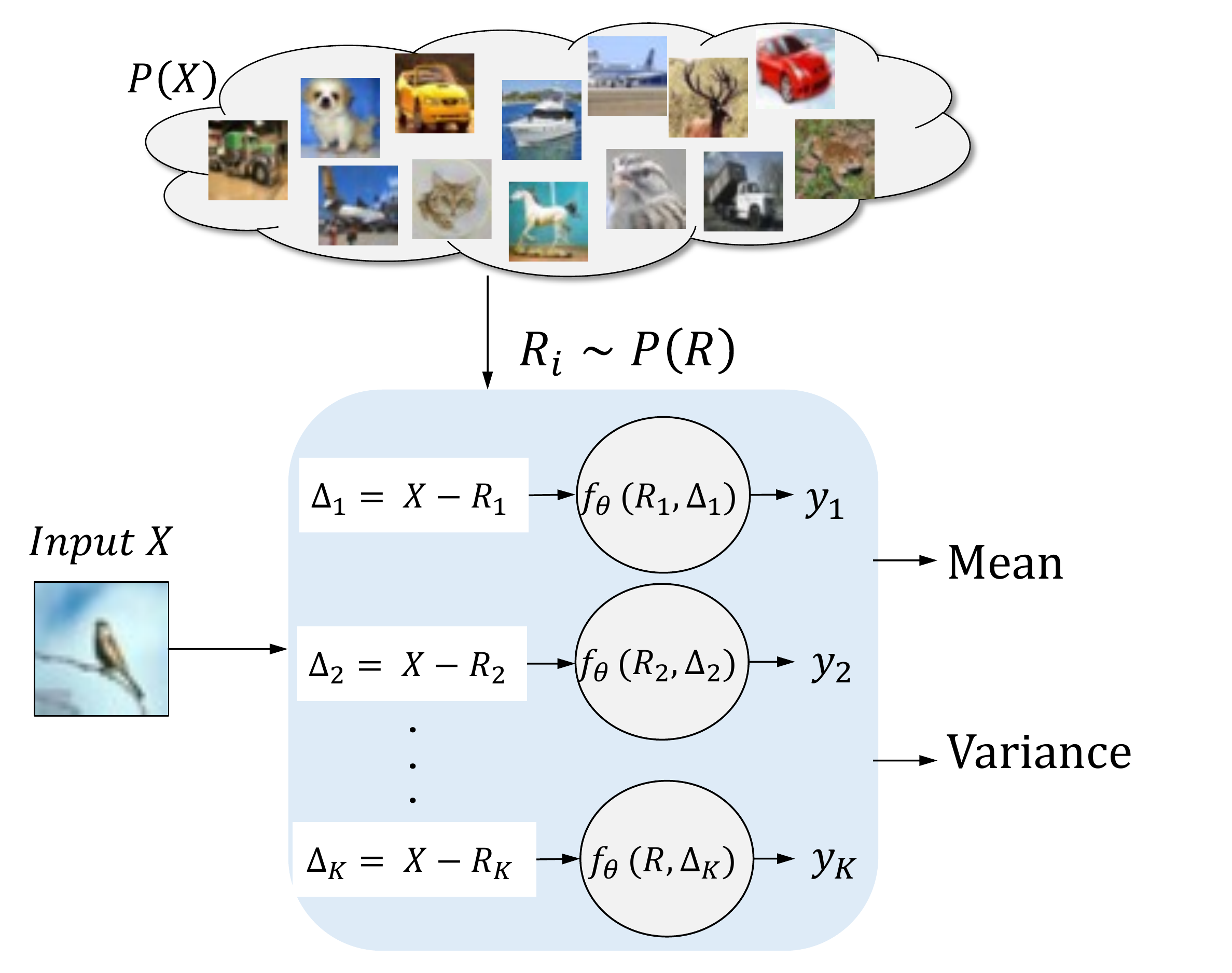}
\label{fig:delta-method}}
\subfloat[][Anchoring in predictive models]{\includegraphics[width=0.4\linewidth,clip,trim=0 0 0 0]{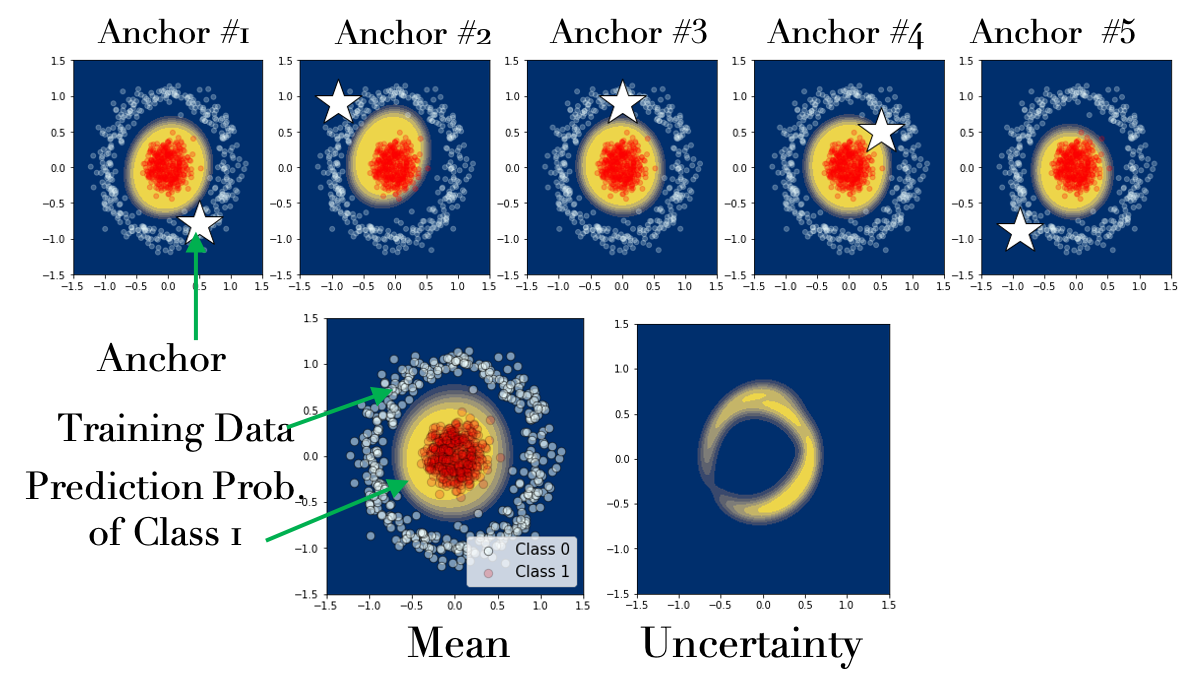}
\label{fig:delta-uq}}
\caption{Illustration of \name.}
\label{fig:teaser}
\end{figure*}

 Figure \ref{fig:delta-method} illustrates our idea of anchoring in predictive models. Interestingly, for a test sample, by marginalizing the effect of the anchor choice from $P(\mathcal{R})$, one can obtain meaningful uncertainty estimates as seen in Figure \ref{fig:delta-uq}. The top row indicates prediction probabilities obtained using our $\Delta-$model for randomly chosen test samples in a $2$D domain (circles denote training data) with $5$ different anchors (indicated as a star), and the bottom row shows the corresponding mean predictions and uncertainty estimates.
\section{Regression Models}

\subsection{Active Learning for Function Optimization}
In the active learning experiment (from Section 4.1.2 of the main paper), we used the following standard $1$D and $2$D functions for evaluation:
\begin{itemize}
    \item \textit{Sinusoid} ($1$D): $f(x) = -\sin(5x^2) - x^4 + 0.3x^3 + 2x^2 + 4.1x$.
    \item \textit{Multi Optima} ($1$D): $f(x) = \sin(x) * \cos(5x) * cos(22x)$.
    \item \textit{Booth} ($2$D): $f([x_1, x_2]) = (x_1 + 2x_2 -7)^2 + (2x_1 +x_2 -5)^2$.
    \item \textit{Levi N.13} ($2$D): $f([x_1, x_2]) = \sin^2(3\pi x_1) + (x_1 -1)^2 [1 + \sin^2(3\pi x_2)] + (x_2 -1)^2 [1 + \sin^2(2\pi x_2)]$.
    \item \textit{Ackley} ($2$D): This standard test function can be defined as follows:
    $f([x_1,x_2]) = -a \exp\left(-b \sqrt{\frac{1}{2} \sum_{i=1}^2 x_i^2}\right) - \\ \exp\left(\sqrt{\frac{1}{2} \sum_{i=1}^2 \cos(cx_i)}\right) + a + \exp(1).$
\end{itemize}

\begin{figure*}[!htb]
\centering
\includegraphics[width=0.8\linewidth,trim={0 0 0 0in}, clip]{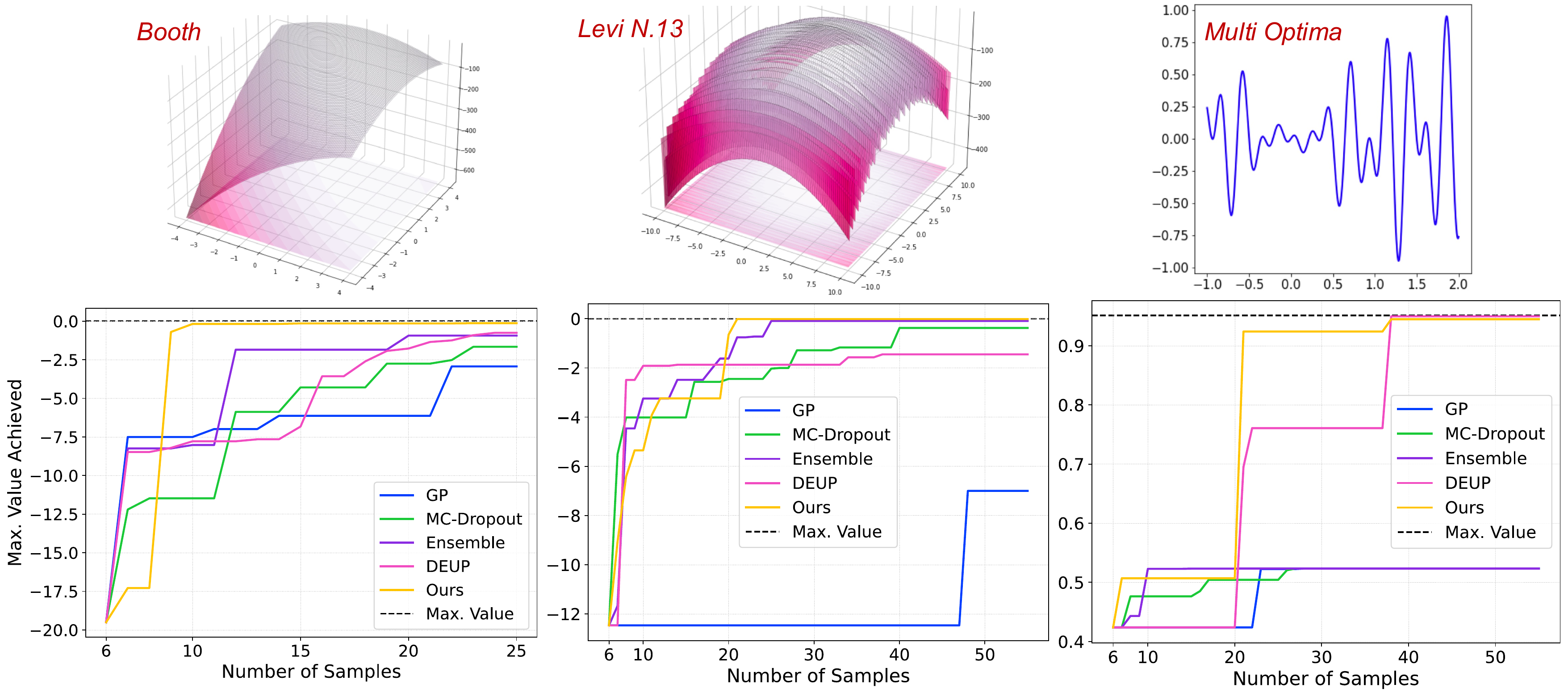}
\caption{\small{Convergence behavior of different uncertainty estimators in an $1-$D sequential optimization. The results reported here correspond to experiments with one random seed and an initial sample size of $6$ samples. We find that \name identifies the true maximum using the least number of samples.}}
\label{fig:smo}
\end{figure*}

\paragraph{Model definition.} For all neural network-based methods, we used the following model architecture: A fully connected network with $3$ hidden layers with $128$ units each and ReLU activation. In the case of deep ensembles baseline, we constructed $M = 3$ independent models with the same architecture (but trained with different random seeds). The MC Dropout baseline used a dropout rate of $0.3$. For implementing DEUP, the error predictor was also implemented as a FCN with $3$ hidden layers ($128$ units). The gaussian process baseline was implemented using the native functions in the BoTorch package.
\paragraph{Optimization.} For each of the test functions and uncertainty estimation method, we started with $6$ samples and incrementally add one sample at a time picked using the \textit{expected improvement} acquisition function. In each iteration, we trained the predictive model using an Adam optimizer with learning rate $1e-3$ for $200$ epochs.

\begin{figure*}[!htb]
     \centering
     \subfloat[Example images synthesized using model-based optimization with uncertainties from the proposed $\Delta-$UQ approach. We show the results for different values of target thickness.]{\includegraphics[width=0.95\linewidth]{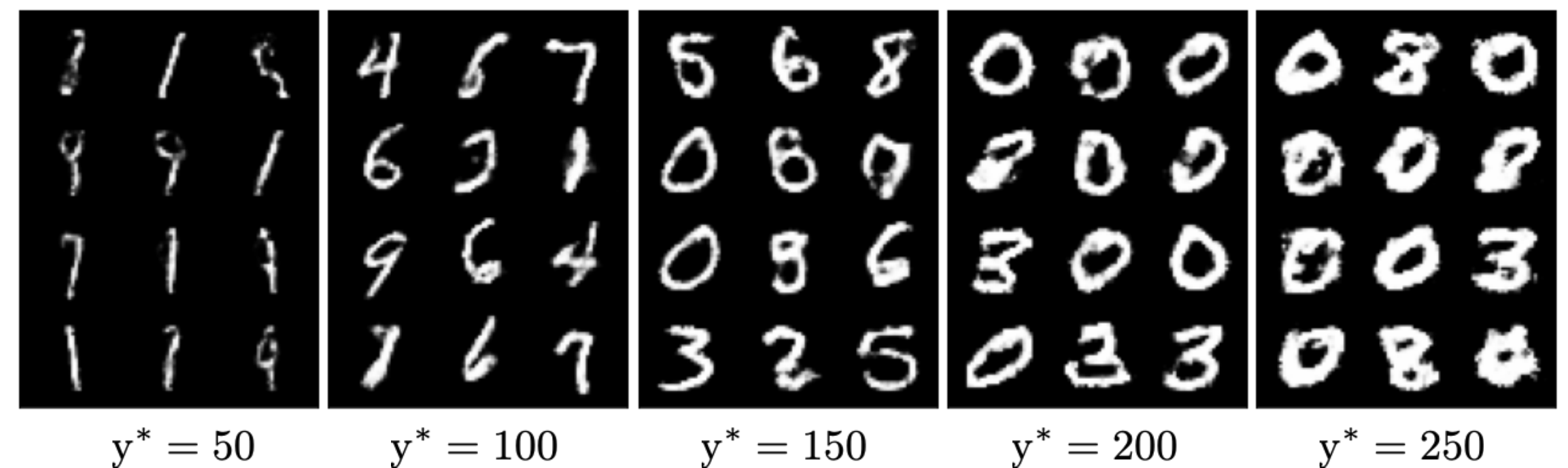}\label{fig:mbo}}

     \subfloat[Number of anchors vs Uncertainties]{\includegraphics[width=0.35\linewidth,clip,trim=0 0 0 0]{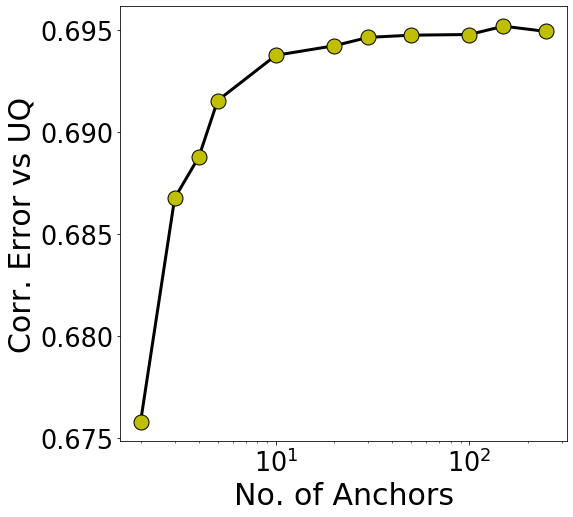}\label{fig:ablation1}}
     \subfloat[Choice of anchor distribution]{\includegraphics[width=0.65\linewidth,clip,trim=0 0 0 0]{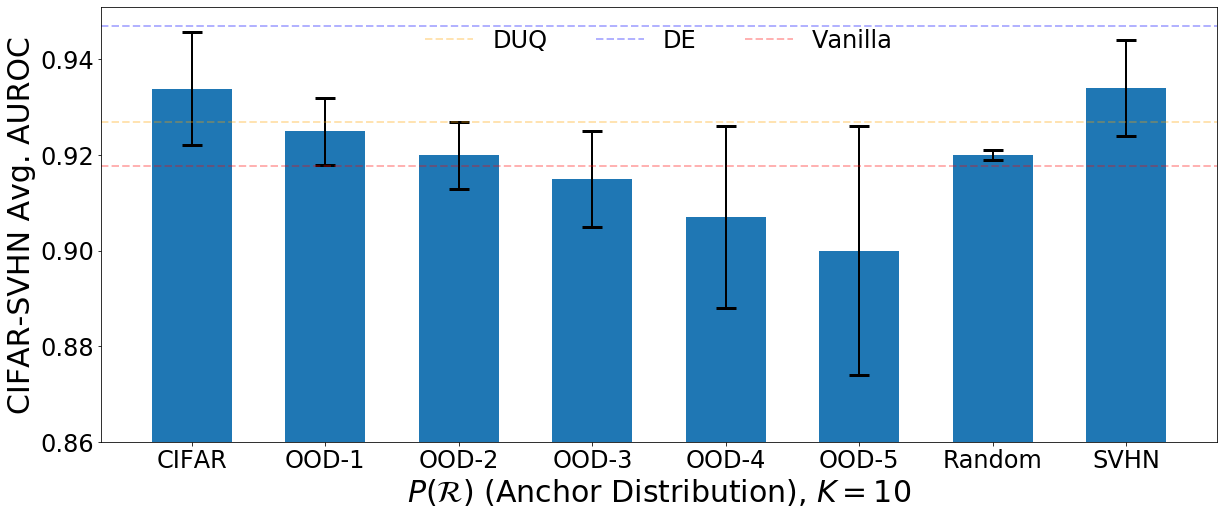}\label{fig:ablation2}}
     \caption{Ablation on Anchor Properties}
     \label{fig:cifarc-ablation}
\end{figure*}
\subsection{Model-based Optimization}
In the model-based optimization experiment (from Section 4.1.3 of the main paper), we considered the MNIST digits dataset and used the digit thickness as the target attribute. Note that, thickness is defined as the total number of pixels with intensity$>0.2$. While the models were trained only using images with thickness$\leq 200$, at test time, we attempt to synthesize images with larger thickness values. Conceptually, we use the uncertainties to run the test-time optimization of searching a relevant solution in the latent space of the inverse model $g_{\phi}$, wherein the loss function is defined solely based on the predictions from the forward model $f_{\theta}$, often referred to as \textit{cyclical} training. By integrating uncertainties into this optimization, one expects to produce solutions that only represent the desired thickness, but can also be characterized by low uncertainty with respect to the forward mode $f_{\theta}$. This will ensure that the identified solutions are reliable and meaningful. We evaluate this behavior by measuring the thickness of the synthesized digits for different values of desired thickness.

\paragraph{Model definition.} The inverse model was fixed to be the same for all methods, while the forward models were separately trained to produce prediction uncertainties. The inverse model $g_{\phi}$ was designed as a FCN using $5$ hidden layers with units $[128, 256, 512, 1024, 784]$, while the input to the inverse model included $100-$dimensional latent noise (with prior $\mathcal{U}[-1,1]$) and the digit thickness as the conditioning variable. All layers (except the last one) used leaky ReLU activation with negative slope $0.2$. The forward model $f_{\theta}$ used the same architecture in all cases and comprised $4$ hidden layers with $512$ units each and ReLU activation.

\paragraph{Optimization.} All models were trained using the Adam optimizer with learning rate $1e-4$ for $100$ epochs. Finally, during testing, we ran the cyclical optimization for $1000$ iterations, wherein for each desired thickness value we used $50$ randomly initialized points in the latent space ($z$) of the inverse model.
Figure \ref{fig:mbo} illustrates example realizations obtained using \name for different values of $y^*$.

\section{Classification Models}

\subsection{Ablation studies on Anchor properties}
First, we study the choice of number of anchors and the distribution from which it is drawn on classification properties. In Figure \ref{fig:ablation1}, we show how the quality of uncertainties estimated using \name change as we vary the number of anchors, $K$, used at test time. As before, we restrict $K=1$ during training. We measure the quality of uncertainties using their correlation between the true generalization error, since it is expected that meaningful uncertainties (both epistemic and aleatoric) are reflected by the generalization error. Here, we choose an out of distribution dataset at test time using CIFAR-10-C (``impulse noise'', intensity 5) to illustrate this aspect. Similar to the regression case in the main text, increasing the number of anchors improves performance until a certain value.

Next, in Figure \ref{fig:ablation2}, we study the impact on performance due to the choice of the anchor distribution, i.e., $P(\mathcal{R})$. We expect an optimal performance when it sufficiently overlaps with the data distribution $P(X)$ and as a result, we use $P(\mathcal{R}) = P(X)$. In this experiment we draw anchors from different prior distributions and measure the impact on performance in the CIFAR-SVHN separation experiment, which uses uncertainties to separate an OOD dataset (SVHN) using  model trained on CIFAR-10. We use the same model as in the main text (ResNet-18). We consider the following different priors: (a) \textbf{OOD:} We choose CIFAR-10-C ("impulse noise") as a third dataset from which we can draw the anchors. By varying the intensity of the corruption, we can systematically measure the performance as the corruptions become more intense (these are labeled ``OOD-1'' in the figure for intensity 1, etc.). (b) \textbf{Random:} Next, we draw anchors from a multi-variate normal distribution with zero mean and unit variance, i.e. $P(\mathcal{R}) = \mathcal{N}(\mathbf{0},\mathbf{I})$, and finally (c) \textbf{SVHN:} we use the SVHN dataset itself to draw random samples as anchors. For all of these cases we use the same distribution for both CIFAR and SVHN datasets, and use $K=10$.

For reference, we also show performance when using anchors from the training distribution itself ("CIFAR"), and for comparison performance against competitive baselines DUQ and Deep Ensembles. We see that choosing CIFAR-10-C as the anchors degrades performance, proportionally to the intensity of the corruption as expected. Interestingly, we see that even a simple random prior distribution gives reasonable performance (better than DUQ), which indicates that a carefully chosen random prior may further improve performance. Finally, choosing anchors from SVHN performs also performs similarly to choosing them from CIFAR itself, indicating the anchors could potentially be used directly at test time itself, without even needing access to the training distribution.
\subsection{Metrics to measure calibration error under distribution shift}
\begin{figure*}[!htb]
     \centering
     \subfloat[Accuracy ($\uparrow$)]{\includegraphics[width=0.99\linewidth,clip,trim=0 0 0 0]{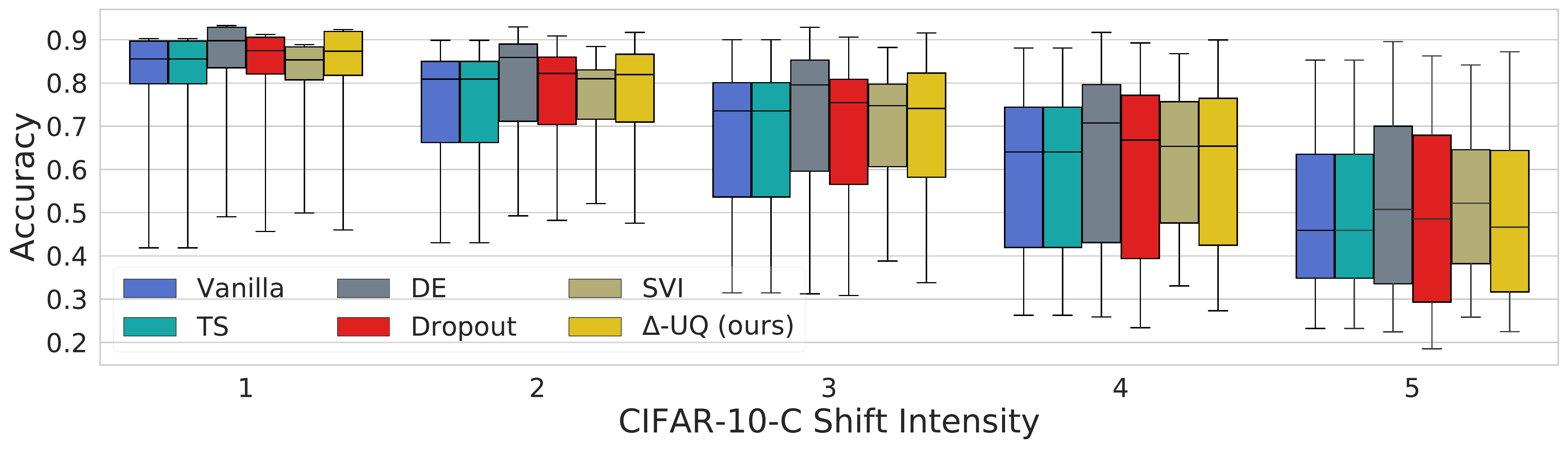}\label{fig:acc}}

     \subfloat[Brier Score ($\downarrow$)]{\includegraphics[width=0.99\linewidth,clip,trim=0 0 0 0]{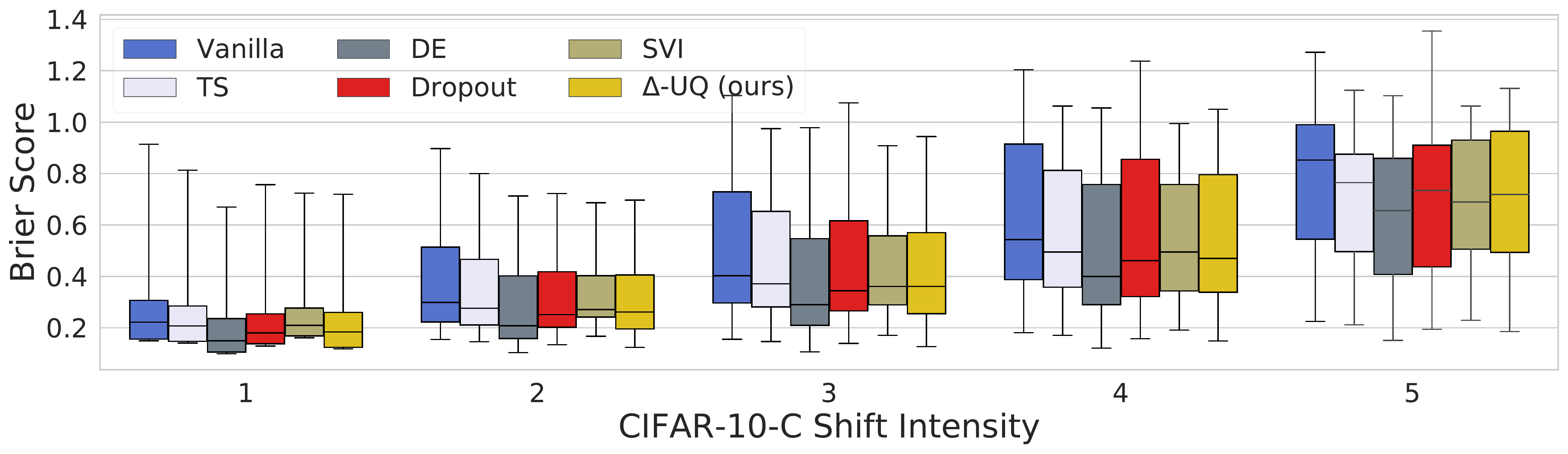}\label{fig:brier}}

     \subfloat[Negative Log-Likelihood ($\downarrow$)]{\includegraphics[width=0.99\linewidth,clip,trim=0 0 0 0]{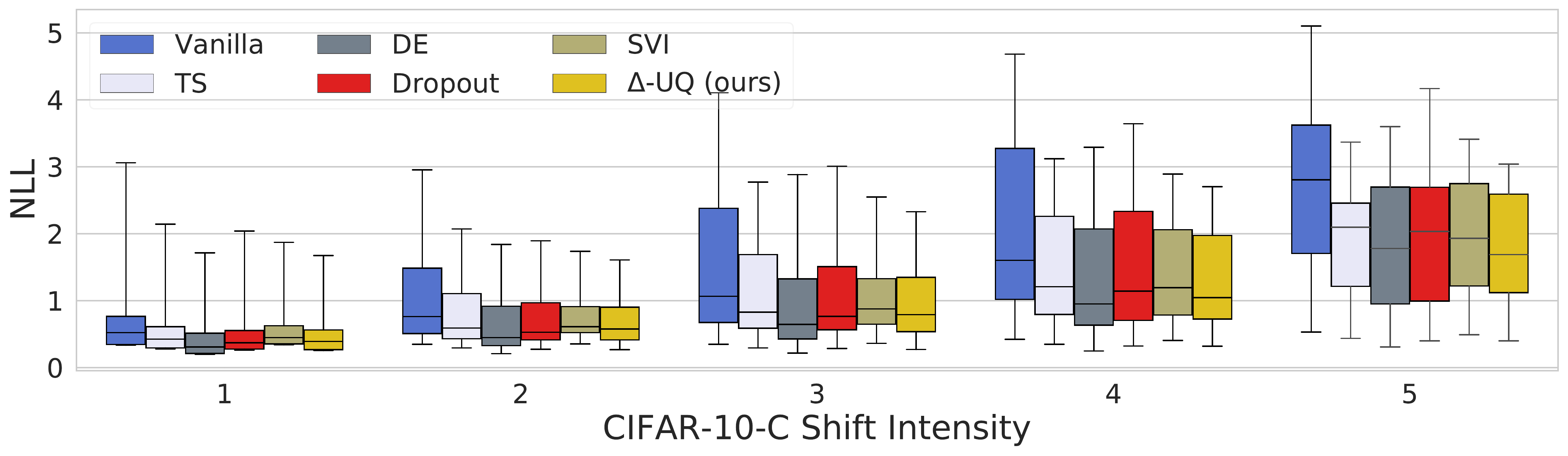}\label{fig:nll}}
     \caption{Measuring accuracy and calibration error under distribution shifts with CIFAR-C.}
     \label{fig:cifarc}
\end{figure*}

In addition to expected calibration error (ECE) used in the main text, here we show additional metrics like Brier Score, and Negative Log Likelihood (NLL) in Figure \ref{fig:cifarc}. We use the same training setup, model, and hyper parameters as in the main text. As before, results are averaged across 5 random seeds. For reference, we also show the accuracy of the model under the different intensities of CIFAR-10-C across all methods. In all cases, we see that the trend observed with ECE holds up, and the proposed method has higher accuracy and lower calibration error compared to most of the single model methods (close to deep ensembles).
\end{appendices}
\end{document}

%% file: intro.tex
Accurately and reliably estimating uncertainties in machine learning (ML) models is arguably as important as advancing predictive capabilities of those models. Uncertainties can inform us about the hardness of a predictive task, or the deficiencies of a model and/or the training data. They also play a crucial role in supporting practical objectives that range from characterizing model behavior under distribution shifts and active learning, to ultimately improving predictive models themselves and building trust. Some of the most popular uncertainty estimation methods today include: (i) Bayesian neural networks~\cite{blundell2015weight,kendall2017uncertainties}: (ii) methods that use the discrepancy between different models as a proxy for uncertainty, such as deep ensembles~\cite{lakshminarayanan2016simple} and Monte-Carlo dropout that approximates Bayesian posteriors on the weight-space of a model~\cite{gal2016dropout}; and approaches that use a single model to estimate uncertainties, such as orthonormal certificates~\cite{tagasovska2018single}, DUQ \cite{van2020uncertainty}, distance awareness~\cite{liu2020simple}, depth uncertainty~\cite{antoran2020depth}, DEUP \cite{jain2021deup} and AVUC \cite{krishnan2020improving}. Despite their demonstrated success in practical tasks, many of these methods are only applicable either to classification settings alone or require special supervised training schemes that may not be broadly applicable. Consequently, an accurate, model-agnostic, single-model uncertainty estimator for both classification and regression problems does not exist today. 

\begin{figure*}[!htb]
	\centering
	\includegraphics[width=0.85\linewidth]{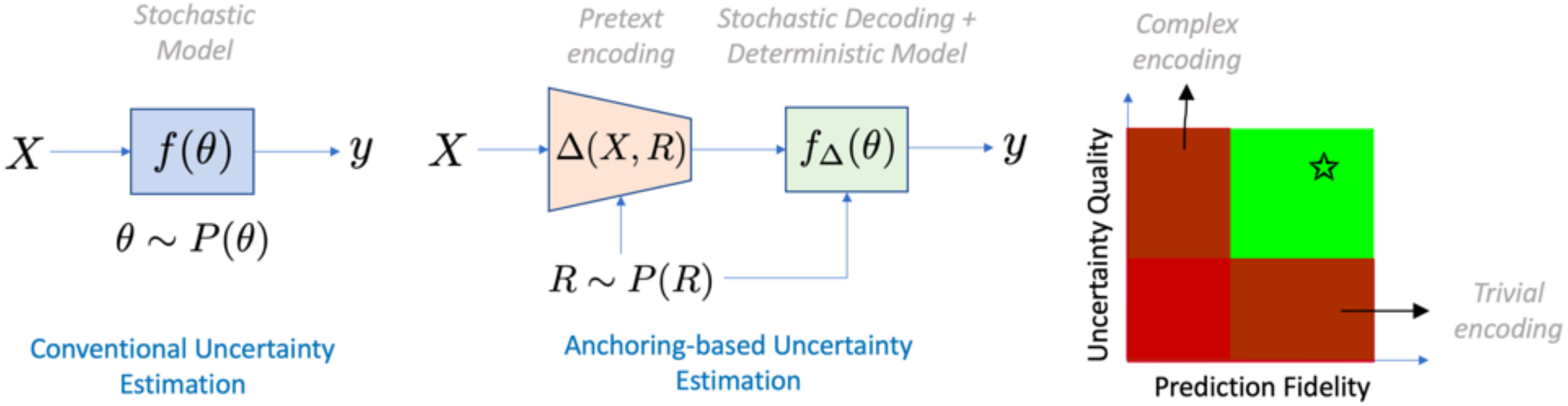}
	\caption{\small{Conventional methods assume the model $f(\theta)$ to be stochastic and measure the model variance as uncertainties. In contrast, the proposed approach implicitly estimates input uncertainties (on $X$) that are propagated through a predictive model. This is achieved using a pre-defined encoding with respect to a reference distribution $P(\mathcal{R})$ and assumes the process of recovering $X$ from $(R, \Delta(X,R))$ to be stochastic, while keeping the predictive model to be deterministic. When the encoding is too simple (for e.g., $\Delta(X,R)=X$) recovering $X$ is trivial, resulting in  poorly calibrated model uncertainties. On the other hand, with very complex encodings the stochastic decoding can produce meaningful uncertainties but can comprise the prediction fidelity. We find that $\Delta(X,R) = X-R$ is just right in terms of high-fidelity predictions as well as calibrated uncertainties.}}
	\label{fig:overview}
\end{figure*}

In this paper, we propose \name-- a general-purpose uncertainty estimator that can be used with any ML model or prediction task. As illustrated in Figure \ref{fig:overview}, our approach first constructs a $\Delta-$model that takes as input a tuple consisting of: (a) A randomly chosen \emph{anchor} point, $R$, drawn from a prior distribution; and (b) a combination of the input, $X$, with $R$ governed by a pretext encoding, $\Delta(X,R)$. This encoding is chosen such that $X$ is recoverable from the tuple $(R, \Delta(X,R))$, for any arbitrary $R$ (for e.g., simple schemes like addition or subtraction can be used). Consequently, the predictive model can be seen as a combination of a \emph{stochastic} decoder--that recovers $X$ given the tuple; and a deterministic predictive model--to predict the target from this estimated input. Once trained, this simple re-parameterization allows us to estimate uncertainties using multiple anchors for a given test sample, by marginalizing out the effect of anchors. We show that these uncertainties are directly linked to improper sampling of the input data distribution and the inherent data noise enabling us to estimate the total uncertainty in the system (combining both aleatoric and epistemic uncertainties), as seen in Figures \ref{fig:overview} and \ref{fig:delta-uq}.


Note, this formulation does not make any assumptions of the kind of model or the task being solved, thus making \name very broadly applicable. In this paper, we show that this strategy of \emph{anchor marginalization} provides a novel framework to understand and capture uncertainties in a system using a single-model. We demonstrate this through extensive empirical studies on a broad range of regression and classification tasks. From improved model fitting, sequential model optimization, and model based inversion in the regression setting, to out of distribution detection and improved model calibration under distribution shifts in classification models. In all these use-cases, we find that \name outperforms many benchmark uncertainty estimators consistently. 


%% file: methods.tex
In this section, we introduce the formulation for $\Delta-$encoding and  describe the proposed approach for anchoring-based uncertainty estimation.

\subsection{Overview}
Let us define a predictive model as $f(\theta): X\rightarrow y$, parameterized by $\theta$, where input $X \in \mathbb{R}^d$ (or $\mathbb{R}^{h \times w \times 3}$ in case of images), and its output $y\in \mathbb{R}^k$. The input $X$ is drawn in an \textit{i.i.d.} fashion from a training distribution $P(\mathcal{X})$ and without loss of generality, $y$ can be continuous-valued or categorical. As illustrated in Figure~\ref{fig:overview}, conventional approaches such as deep ensembles assume the model $f(\theta)$ to be stochastic and utilize the model variance as uncertainties. In contrast, we define a pretext encoding as function-- $\Delta(X,R)$, where $R$ denotes an anchor of the same dimensions as $X$, drawn from a prior distribution $P(\mathcal{R})$. Here on, we denote the pretext task as $\Delta_{X,R}$ for conciseness. We repose the learning task as $f_\Delta(\theta): \{R, \Delta_{X,R}\} \rightarrow y$, where the input is a tuple obtained by concatenation when $X$ is a vector or appending along channels when $X$ is an image. Since $y$ is only dependent on $X$, $R$ behaves as a distractor to the model $f_\Delta(\theta)$. Consequently, our approach assumes a stochastic decoding process (w.r.t. choice of $R$), while the predictive model is itself deterministic.

\paragraph{} The central idea behind \name is to exploit the inconsistencies of the model in decoding $X$, when it is tasked with predicting the target given only the tuple $\{R, \Delta_{X,R}\}$. For a more intuitive exposition, we consider an ``explicit'' formulation that separates the tasks of recovering $X$ and predicting $y$ from $\{R, \Delta_{X,R}\}$. First, let $g:\{R,\Delta_{X,R}\}\rightarrow X$, followed by $h: X \rightarrow y$. Considering $g$ is stochastic ($R$ is randomly drawn for a fixed $X$), the uncertainties in recovering $X$ are propagated through $h$, to obtain uncertainties on the target variable $y$. Interestingly, due to the non-linear nature of the predictor $h$, the uncertainties from $g$ affect both the mean and variance estimates of $h$~\cite{gast2018lightweight}. At inference time, one can marginalize the choice of $R$, by obtaining multiple realizations of $f_{\Delta}^{\text{explicit}} = h \circ g (R, \Delta_{X,R})$  and measuring the variance. Uncertainties in recovering $X$ can arise from insufficient sampling in training data (epistemic) or inherent noise in $X$ (aleatoric). We make a few observations:
\begin{itemize}
 \setlength{\itemindent}{0em}
    \item When $g$ learns to exactly recover $X$, there are no uncertainties in the decoding since this reduces exactly to a standard predictive model.
    \item When $X \sim P(\mathcal{X})$ is not properly sampled -- as is the case in most high-dimensional distributions -- poorly sampled regions can lead to larger discrepancies in recovering $X$ under different choices of $R \sim P(\mathcal{R})$, and thereby larger prediction uncertainties in $y$.
\end{itemize}In practice, we combine the  $g$ and $h$ models into a single model $f_\Delta$, where the model can determine the level of error tolerance in the decoding process.

\subsection{Anchoring-based Uncertainty Estimation}
An anchor $R$ is a sample defined in the $d-$dimensional input domain $\mathcal{X}$ that can be used as a reference to make inferences about an input sample $X$. We posit that, for a reliable prediction, one must be able to decode $X$ from its pretext encoding, regardless of the choice of the anchor $R$. For simplicity, we assume the prior distribution for the anchors $P(\mathcal{R}) = P(\mathcal{X})$, though it is not required. In this paper, we show how the principle of anchoring can be used to design accurate uncertainty estimators for any arbitrary machine learning model and both continuous and categorical-valued prediction tasks. In order to perform anchoring-based uncertainty estimation, we construct the model $f_{\Delta}$ to take the encoding $\Delta_{X,R}$ as an additional input along with the anchor $R$, and ensure that the model's prediction $y$ is consistent with any anchor $R \sim P(\mathcal{R})$. 

\paragraph{Training and inference} During training, we use a randomly chosen anchor for every input sample over multiple epochs so that each input sample is combined with different anchors over the entire training process. At inference time, we estimate the prediction uncertainty for a test sample by analyzing the consistency in predictions with respect to $K-$different anchors. In other words, for a given test sample $X$, we marginalize the effect of choosing an anchor and compute the mean prediction along with its uncertainty as follows:
\begin{align}
\label{eq:delta_uq}
\mu_X &= \frac{1}{K} \sum_k f_{\Delta}(R_k, \Delta_{X, R_k}; \theta),\\
\sigma_X^2 &= \frac{1}{K}\sum_k\left[f_{\Delta}(R_k, \Delta_{X, R_k}; \theta)-\mu_X\right]^2
\end{align}where $R_k\sim P(\mathcal{R})$. 

\paragraph{Choice of pretext encoding $\Delta$.}
The quality of uncertainties in our approach relies directly on the complexity of $\Delta$. When we define a trivial encoding $\Delta_{X,R} = X$, wherein the decoding process has to ignore $R$ when presented the tuple $\{R,X\}$. Most ML models can easily achieve this and hence produce a deterministic model with unreliable or trivial uncertainties. As a more complex alternative, we consider the encoding $\Delta_{X,R} = X-R$, where the decoding process is to compute the sum $R + \Delta_{X,R}$ for any $R \in P(\mathcal{R})$. To understand the impact of different design choices for $\Delta$ encoding, we study regression experiments with standard benchmark functions and a decision tree model, as shown in Table \ref{tab:delta}. We measure both the prediction fidelity (R2 statistic) and the calibration (Spearman correlation between uncertainties and prediction error). As discussed earlier, with the na\"ive encoding of $\Delta_{X,R} = X$, the resulting uncertainties are meaningless (indicated by low Spearman correlation). On the other hand, adopting a high-complexity encoding with two anchors, i.e., $\Delta_{X, R_1, R_2} = X - R_1 - R_2$, leads to well-calibrated uncertainties but with lower fidelity mean estimates. In comparison, the encoding $\Delta_{X, R} = X - R$ performs the best in terms of both metrics and induces the appropriate level of complexity. For all experiments reported in the rest of this paper, we used this pretext encoding.
\begin{table}[htb]
	\centering
	\small{
	\caption{\small{Choice of pretext encoding for $\Delta-$UQ. Using standard regression functions in different dimensions and a decision tree regressor, we study the impact of increasing encoding complexity on the prediction fidelity and uncertainty calibration. For each case, we show the R2 statistic / spearman correlation between uncertainties and generalization error.}}
	\vspace{-5pt}

		\renewcommand*{\arraystretch}{1.5}
	\begin{tabular}{l|c|c|c}

		\hline
		\rowcolor[HTML]{C0C0C0} 
		\multicolumn{1}{c|}{\cellcolor[HTML]{C0C0C0}} & \multicolumn{3}{c}{\cellcolor[HTML]{C0C0C0}\textbf{Choice of $\Delta$}~~~~ \small{R2($\uparrow$) / Corr. ($\uparrow$)}} \\
		\cline{2-4}
		\rowcolor[HTML]{EFEFEF} 
		\multicolumn{1}{c|}{\multirow{-2}{*}{\cellcolor[HTML]{C0C0C0}\textbf{Function}}} & \textbf{$X$} (na\"ive)        & \textbf{$X-R$}        & \textbf{$X - R_1 - R_2$}       \\
		\hline \hline
		Ackley-2D                                                                        & 0.83 / 0.09                & 0.85 / 0.31                  & 0.81 / 0.35                           \\
		Ackley-3D                                                                        & 0.81 / 0.04                & 0.84 / 0.34                  & 0.79 / 0.38                           \\
		Ackley-4D                                                                        & 0.77 / 0.07                & 0.81 / 0.35                  &0.76 / 0.41                          \\
		\hline \hline
		Griewank-2D                                                                        &0.99 /  0.1                & 0.99 / 0.49                  & 0.98 / 0.51                           \\
		Griewank-3D                                                                        & 0.97 / 0.09                & 0.98 / 0.52                  & 0.96 / 0.56                           \\
		Griewank-4D                                                                        & 0.95 / 0.12                & 0.96 / 0.53                  & 0.94 / 0.55                          \\
		
		\hline            
	\end{tabular}
\label{tab:delta}}
\end{table}
\begin{figure}[!t]
	\centering
   \includegraphics[width=0.95\linewidth,clip,trim=0 0 0 0]{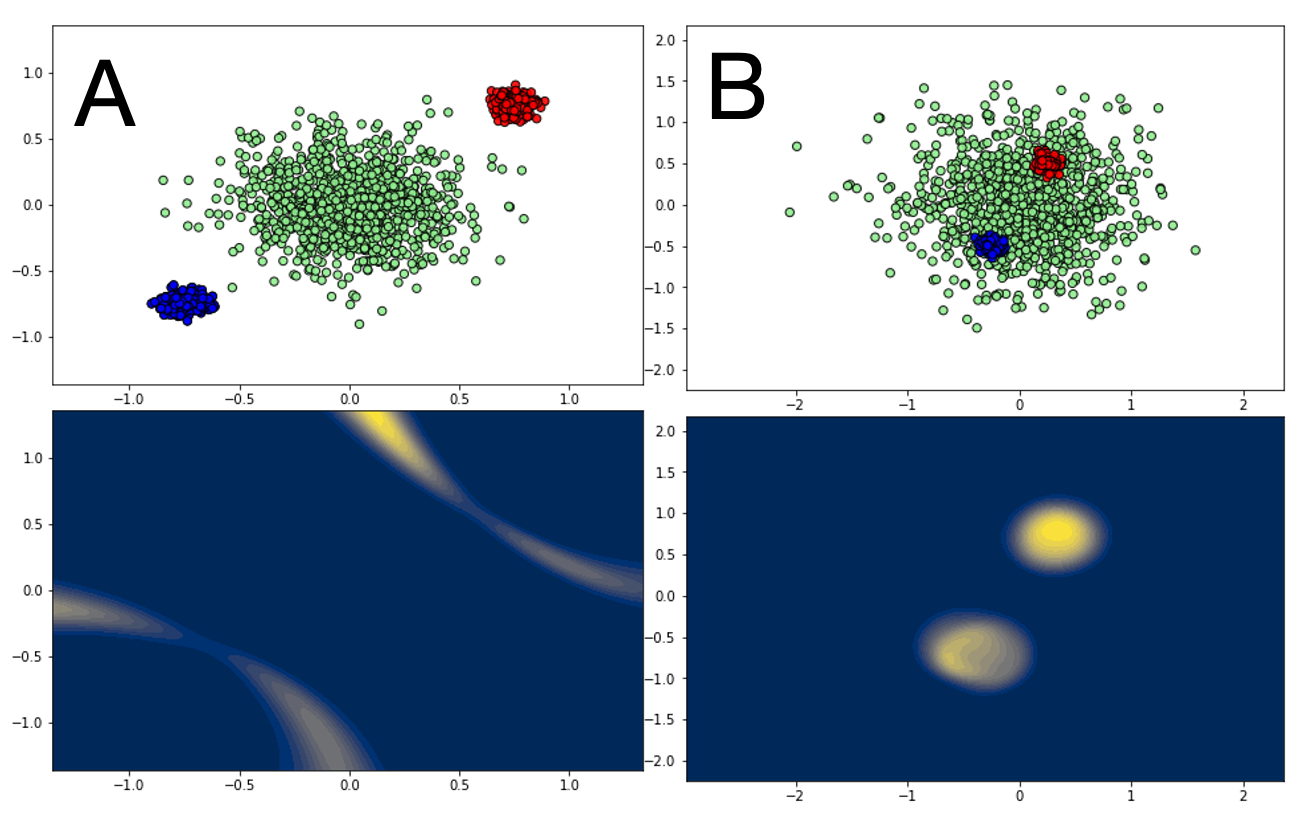}
	\caption{\small{\textbf{Uncertainty types:} In the bottom row we demonstrate uncertainties obtained using \name in a toy classification problem. The top row shows the two scenarios -- (A) where the largest uncertainties are mostly along the decision boundaries (\emph{epistemic}); and in (B) the two smaller classes are nearly inseparable from the larger, green class, where the uncertainties come from lack of sufficient knowledge (\emph{aleatoric}). Here the uncertainty is obtained as the sum of variance in prediction probabilities of all three classes of the model.}}
	\label{fig:delta-uq}
	\vspace{-5pt}

\end{figure}

\paragraph{Demonstration.} In Figure \ref{fig:delta-uq}, we illustrate the behavior of \name using a simple $3$-way classification setup in two dimensions with a kernel SVM model~\cite{cortes1995support}. Once trained, we define the test data as a uniform sampling of the entire $2D$ domain (assuming that the bounds are known), and obtain predictions for each test sample using $5$ randomly chosen anchors. The bottom row of Figure \ref{fig:delta-uq} shows the total uncertainty of the model -- measured by the sum of variance in prediction probabilities of all three classes obtained using the $\Delta-$kernel-SVM. We observe that when the classes are perfectly separable (Fig \ref{fig:delta-uq}A), the largest uncertainties arise from the lack of training points near the class boundaries. On the hand, when the blue and red classes are nearly inseparable from the green class (Fig \ref{fig:delta-uq}B), even though they are densely sampled, \name effectively reflects the aleatoric uncertainties associated with those two classes.

\begin{table*}[!htb]
\centering
\caption{\small{Benchmarking \name on regression tasks from the OpenML dataset collection. We observe that our approach consistently out performs competitive baselines.}}
\renewcommand{\arraystretch}{1.3}
\begin{tabular}{c|c|c|c|c|c|c}
\hline
\rowcolor[HTML]{C0C0C0} 
\cellcolor[HTML]{C0C0C0}     & \multicolumn{3}{c|}{\cellcolor[HTML]{C0C0C0}Spearman Correlation $\uparrow$}  & \multicolumn{3}{c}{\cellcolor[HTML]{C0C0C0}MAE $\downarrow$} \\ \cline{2-7} 
\rowcolor[HTML]{EFEFEF} 
\multirow{-2}{*}{\cellcolor[HTML]{C0C0C0}Dataset ID} & \multicolumn{1}{l|}{\cellcolor[HTML]{EFEFEF} Ensemble } & \multicolumn{1}{l|}{\cellcolor[HTML]{EFEFEF}DEUP} & \multicolumn{1}{l|}{\cellcolor[HTML]{EFEFEF}\name} & \multicolumn{1}{l|}{\cellcolor[HTML]{EFEFEF}Ensemble} & \multicolumn{1}{l|}{\cellcolor[HTML]{EFEFEF}DEUP} & \multicolumn{1}{l}{\cellcolor[HTML]{EFEFEF}\name} \\ \hline \hline
201 & 0.764 & 0.510 & \textbf{0.867} & 13.45 & 14.21 & \textbf{12.29} \\ 
216 & {0.41} & 0.40 & \textbf{0.51} & 0.004 & 0.004  & \textbf{0.003} \\ 
296 & 0.201 & \textbf{0.281} & 0.266 & \textbf{0.0002}  & \textbf{0.0002}  & \textbf{0.0002}  \\ 
1199  & 0.265  & 0.123     & \textbf{0.283} & 9.87  & 10.14     & \textbf{9.34}   \\ 
23515 & 0.13    & \textbf{0.287}     & \textbf{0.285}   & \textbf{0.02}  & \textbf{0.02}     & \textbf{0.02}   \\ 
41539 & 0.594    & 0.411     & \textbf{0.665}   & \textbf{110.88}   & 119.59    & 118.91     \\ 
42193 & \textbf{0.459}    & 0.239     & 0.439 & 3.07     & 2.99 & \textbf{2.97}    \\ 
42688 & 0.639    & \textbf{0.674} & 0.642 & 892.03  & 903.57    & \textbf{705.94}  \\ 
42712 & \textbf{0.634}    & 0.547     & 0.598 & 82.77    & 83.21     & \textbf{81.76}   \\\hline
\end{tabular}
\label{table:reg}
\end{table*}

%% file: related.tex
Bayesian methods \cite{neal2012bayesian} are among the most common kinds of uncertainty estimators today, but they are not easily scalable to large datasets and are known to be outperformed by ensemble methods \cite{lakshminarayanan2016simple}. Monte Carlo Dropout \cite{gal2016dropout} is a scalable alternative to Bayesian methods in that it approximates the posterior distribution on the weights via dropout to estimate uncertainties. However, it is well known that dropout tends to over estimate uncertainties due to sampling the parameter space in regions that are potentially far away from the local minima. Deep Ensembles \cite{lakshminarayanan2016simple} showed a simple way to obtain accurate uncertainties in a task agnostic and model agnostic fashion. While extremely accurate and currently one of the best uncertainty estimation techniques \cite{Ovadia2019}, it requires training several models with different random initializations which can become a computational bottleneck when training deep networks. 

As a result, there have been several methods that seek to obtain accurate uncertainties from just a single model, typically a deep neural network. DUQ \cite{van2020uncertainty} uses a kernel distance to a set of class-specific centroids defined in the feature of a deep network as the measure for uncertainty. Another recent technique is DEUP \cite{jain2021deup} which trains an explicit epistemic uncertainty estimator for a pre-trained model, which can also be used efficiently in a task-agnostic manner. In contrast, \name does not ivolve any post-hoc step and estimates the inherent uncertainties via the process of anchor marginalization.



%% file: expts.tex
We benchmark the quality of uncertainties estimated using \name in this section. Since it is a general purpose estimator, we evaluate \name in a range of diverse tasks: model fitting, sequential model optimization, and model based inversion in regression; out of distribution detection and calibration under distribution shifts for classification. 

\subsection{Regression Models}
\subsubsection{Uncertainty Calibration on OpenML Benchmarks}
We validate the quality of the uncertainty estimates from $\Delta-$models using standard OpenML \footnote{\url{https://www.openml.org/search?type=data}} regression tasks. Table \ref{table:reg} lists the dataset IDs considered in this experiment. 

\paragraph{\emph{Experimental setup}}
In this study, we build a random forests regressor (with $5$ trees) for each of the datasets. Note that, in each case, we fixed the number of training samples at $200$ and used the remaining data samples for testing. We evaluate the performance using two important metrics: (i) mean absolute error (MAE) for the mean prediction; and (ii) Spearman rank correlation between the prediction error and the estimated uncertainties, similar to~\cite{jain2021deup}. Preserving the ranking for the entire test set (including in-distribution samples) is a difficult task for any uncertainty estimator, and the limited training data makes it even more challenging. We repeated the experiments for $5$ independent trials (different random seeds and train-test splits). For comparison, we show results for an ensemble of random forest regressors (ensemble size $M = 3$) and the more recent direct uncertainty prediction approach (DEUP)~\cite{jain2021deup}. Note that, for DEUP, we used an additional validation set of $100$ samples to train the error predictor. We fixed the number of anchors for \name at $100$.

\paragraph{\emph{Results}} Table \ref{table:reg} reports the average performance obtained from $5$ independent trials. We observe that our approach consistently produces higher quality uncertainty estimators in all cases, as indicated by the higher Spearman rank correlation values. While existing uncertainty estimation strategies such as ensembling and DEUP show a large variance in the correlation values across different datasets, \name does not have that behavior. This can be attributed to the fact that, even with a smaller training set, our approach allows re-parameterization of each sample w.r.t. different anchors and thereby produces meaningful uncertainties in the stochastic decoding process.

\begin{table*}[!t]
\centering
\small{
\caption{\small{\textbf{Sequential model optimization (SMO):} Performance comparison of different uncertainty estimation strategies in SMO. In each case, we show the maximum value reached upon active sample collection for the same number of iterations, based on \textit{expected improvement} acquisition function. The results reported were obtained using $5$ independent trials with different random seeds.}}
\renewcommand{\arraystretch}{1.3}
\begin{tabular}{l|c|c|c|c|c}
\hline
\rowcolor[HTML]{EFEFEF} 
\multicolumn{1}{c|}{\cellcolor[HTML]{C0C0C0}}      & Booth      & Levi N.13    & Multi Optima    & Ackley      & Sinusoid     \\  
\rowcolor[HTML]{EFEFEF} 
\multicolumn{1}{c|}{\multirow{-2}{*}{\cellcolor[HTML]{C0C0C0}Method}} & (Max = 0)    & (Max = 0)      & (Max = $0.951$)   & (Max = 0)     & (Max = $7.622$)  \\ \hline \hline
GP     & -0.13 $\pm$ 0.05 & -4.2 $\pm$2.27 & 0.79 $\pm$0.16  & \textbf{-0.09 $\pm$0.04} & \textbf{7.622 $\pm$ 1e-4} \\ 
MC-Dropout \shortcite{gal2016dropout} & -0.58 $\pm$ 0.31 & -0.62 $\pm$0.27     & 0.82 $\pm$ 0.11 & -0.29 $\pm$ 0.17  & 7.37 $\pm$0.46 \\ 
Deep Ens. \shortcite{lakshminarayanan2016simple}     & -0.41 $\pm$ 0.39 & -0.36 $\pm$0.41     & 0.74 $\pm$0.14  & -0.24 $\pm$ 0.21  & 7.59$\pm$0.04  \\ 
DEUP \shortcite{jain2021deup}   & -0.31$\pm$0.36    & -0.16 $\pm$0.12     & 0.85 $\pm$0.08  & -0.18 $\pm$0.14    & \textbf{7.622$\pm$2e-4}   \\ 
\name (ours)     & \textbf{-0.08$\pm$0.09} & \textbf{-0.06 $\pm$ 0.15} & \textbf{0.89 $\pm$0.09}  & -0.14 $\pm$0.08    & \textbf{7.622$\pm$1e-4}   \\ \hline
\end{tabular}
\label{table:smo}
}
\end{table*}

\subsubsection{Active Learning for Function Optimization}
Active learning is an important evaluation task for uncertainty estimators, and its primary objective is to determine which samples to query in order to improve a model. In general, when such sequential optimization techniques are used to determine the maximum (or minimum) value of the function that the model is trying to approximate, it is common to choose samples based on both the predicted value and the uncertainties. Commonly referred to as the \textit{exploration-exploitation} trade-off, this approach defines an acquisition function $\mathcal{A}$ (\textit{e.g.}, expected improvement~\cite{movckus1975bayesian} or upper confidence bound~\cite{srinivas2009gaussian}) and chooses the input $X$ that maximizes $\mathcal{A}$. In this experiment, we consider standard optimization functions in $1$D and $2$D and evaluate the effectiveness of \name estimates in performing sequential model optimization.

\begin{figure}[!t]
\centering
\includegraphics[width=1\linewidth,trim={0 0 0 0in}, clip]{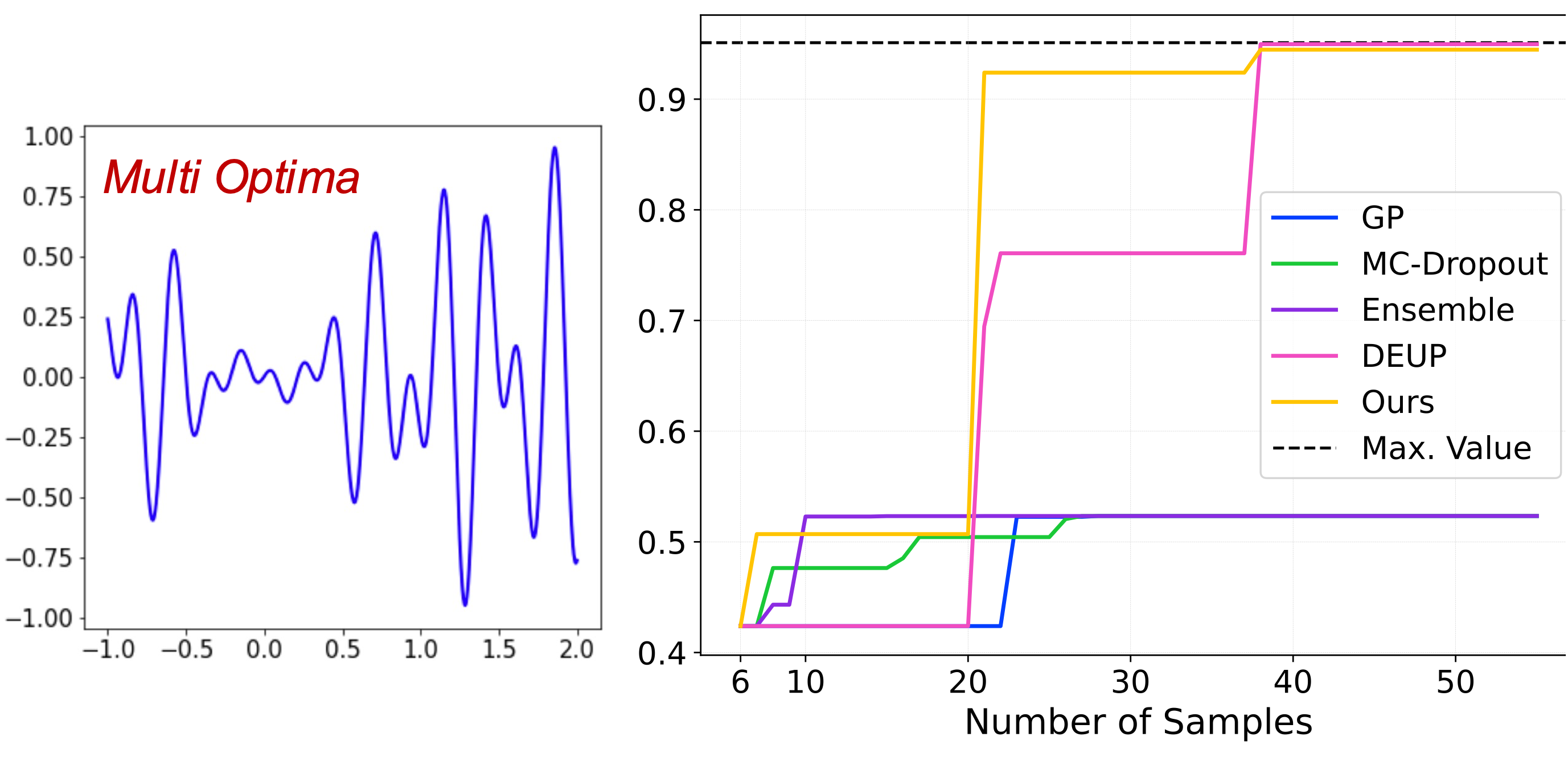}
\caption{\small{Convergence behavior of different uncertainty estimators in an $1-$D sequential optimization. The results reported here correspond to experiments with one random seed and an initial sample size of $6$ samples. We find that \name identifies the true maximum using the least number of samples. Results for more functions can be found in the supplement.}}
\label{fig:smo}
\end{figure}

\paragraph{\emph{Experimental setup}} For this study, we use the following functions: (a) Booth; (b) Levi N.13; (c) Ackley; (d) Sinusoid; and (e) Synthetic function with multiple optima in $1$D (Multi-Optima, see Figure \ref{fig:smo}). In each of the cases, we start with an initial sample size of $6$ samples drawn randomly from an uniform distribution, and sequentially include one sample in each iteration. The regressor model is implemented as a deep neural network with ReLU activation. For comparison, we used the following methods: (i) Gaussian Processes (GP); (ii) MC Dropout with dropout rate $0.3$; (iii) deep ensembles with $3$ models; and (iv) DEUP~\cite{jain2021deup}. In all the neural network based methods, including $\Delta$-UQ, we treat the mean and uncertainty estimators as the mean and variance of a Gaussian distribution in order to compute the acquisition function (expected improvement). We used the BoTorch\footnote{\url{https://botorch.org/}} package for all our experiments. 

\paragraph{\emph{Results}} Figure~\ref{fig:smo} illustrates the results for an $1-$D multi-optima function after $50$ iterations (for one random seed). In the case of Booth (see Supplement material), while all baselines including GP reach close to the maximum value, \name requires lesser number of iterations to reach the maximum value. However, with Levi N.13 (see Supplement material), only the neural network surrogates are able to effectively reach the global maximum. In particular, deep ensembles and $\Delta$-UQ perform better than the other methods, both in terms of the max. value achieved as well as the number of iterations required. Finally, with the more complex Multi-optima function (Figure~\ref{fig:smo}), DEUP is the only other baseline reaching the maximum ($\approx 0.951$), albeit requiring twice the number of iterations compared to \name. Table \ref{table:smo} shows the results for all functions, \textit{i.e.}, the best value achieved after the specified number of iterations ($20$ for Booth, $50$ for the rest), averaged across $5$ random seeds. We see that, \name consistently produces higher-quality uncertainty estimates and outperforms existing approaches.

\begin{table*}[t]
\centering
\caption{\small{\textbf{Model-based optimization (MBO):} Performance comparison of different uncertainty estimators for MBO. Using an approach similar to~\cite{kumar2019model}, we synthesize inputs for a given target thickness with pre-trained $f_{\theta}$ and $g_{\phi}$ models.}}
\renewcommand{\arraystretch}{1.3}
\begin{tabular}{l|c|c|c|c|c}
\hline
\rowcolor[HTML]{EFEFEF} 
\multicolumn{1}{c|}{\cellcolor[HTML]{C0C0C0}{Target Thickness}} & {150} & {175}    & {200}    & {225}  & {250} \\ \hline \hline
Vanilla \shortcite{kumar2019model} (no unc.)  & 143.8 $\pm$ 2.2 & 167.97 $\pm$ 2.8 & 191.2 $\pm$ 3.55 & 215.9 $\pm$ 4.2    & 238.8 $\pm$ 7.0     \\ 
MC Dropout \shortcite{gal2016dropout}  & 143.2 $\pm$ 1.9  & 167.5 $\pm$ 3.1 & 190.9 $\pm$ 2.4 & 214.3 $\pm$ 3.9    & 236.9 $\pm$ 4.7    \\ 
Deep Ens. \shortcite{lakshminarayanan2016simple}   & 146.1 $\pm$ 2.1  & 169.1 $\pm$ 3.7 & 192.3 $\pm$ 1.9 & 219.4 $\pm$ 4.2  & 239.4 $\pm$ 5.9   \\ 
\name (ours)     & \textbf{148.6 $\pm$ 2.3} & \textbf{172.8 $\pm$ 3.1} & \textbf{196.8 $\pm$ 2.4}  & \textbf{222.7 $\pm$ 2.6} & \textbf{244.3 $\pm$ 5.1}  \\ \hline
\end{tabular}
\label{tab:mbo}
\end{table*}

\subsubsection{Model-Based Optimization}
In this experiment, we consider the problem of model-based optimization, wherein one needs to estimate the unknown high-dimensional input $X$ for a given target $y$ using a predictive model trained on offline data (without active data collection). Commonly referred as an inverse model, the goal is to train a mapping $g_{\phi}: y \mapsto X$ using labeled data and query $g_{\phi}$ at test-time to synthesize relevant samples. Given that, multiple $X$ can map to the same target $y$, it is common to include additional latent variables $z$ (with a known prior $P_z$) to implement $g_{\phi}: [y,z] \mapsto X$. While a large family of generative models exist for approximating such conditional distributions $P(X|y)$, it was recently found in~\cite{kumar2019model}, one can improve the quality of such inverse models by leveraging a corresponding forward map $f_{\theta}:X \mapsto y$ during test-time optimization. Formally, for a given target $y^*$, this optimization can be written as:
$$
z^* = \arg \min_z |f_{\theta}(g_{\phi}([y^*,z])) - y^*| ~~\text{ s.t. }~~ \log P_z(z) \geq \epsilon.
$$While this has been showed to be superior to using only the inverse $g_{\phi}$, we propose to augment prediction uncertainties into this optimization. More specifically, we will modify this formulation to promote solutions $z$ that not just produce the desired value $y^*$ when we compute $f_{\theta}(g_{\phi}([y^*,z])$, but can also be characterized by low uncertainties. To this end, we view the predicted mean and uncertainties from $f_{\theta}^{\Delta}$ as the location ($\mu$) and scale ($b$) parameters of a Laplacian distribution and solve
\begin{multline}
z^* = \arg \min_z |f_{\theta}(g_{\phi}([y^*,z])) - y^*| \exp(-v) + v \\~~\text{ s.t. }~~ \log P_z(z) \geq \epsilon,~~v = \log b.    
\end{multline}
\paragraph{\emph{Experimental Setup}}For this study, we consider MNIST handwritten digits and define the \textit{digit thickness} as the target, \textit{i.e.}, the total number of pixels with intensity greater than $0.2$. While $f_{\theta}$ is implemented as a fully connected network, $g_{\theta}$ is a CNN model with a $100-$dimensional noise latent space ($z$) with a uniform prior $\mathcal{U}[-1,1]$. We trained both the models using images with thickness lower than $200$, though we used higher values for $y^*$ at test time. 

\paragraph{\emph{Results}} Table \ref{tab:mbo} shows the performance of different uncertainty estimators in model-based optimization. More specifically, we show the thickness of the synthesized images (averaged across $100$ realizations) for different values of $y^*=[150, 175, 200, 225, 250]$. Our results show that \name consistently produces $X^*$ that meets the desired target, while also outperforming MC Dropout and deep ensembles. This clearly demonstrates the ability of our approach to produce meaningful uncertainties even for samples outside the training regimes (thickness$> 200$). We show example realizations obtained using \name in the supplement.

\subsection{Classification Models}
\subsubsection{Uncertainty based OOD Detection}

Here, we validate the estimated uncertainties by using them to detect out of distribution (OOD) samples, based on the fact that we expect OOD samples to have significantly larger uncertainties than those similar to the training distribution. We demonstrate this using a ResNet-18 \cite{he2016deep} model trained on CIFAR-10 \cite{krizhevsky2014cifar}, to detect SVHN \cite{svhn} as OOD due to both being images of size $32\times 32$.

\paragraph{\emph{Experimental Setup}} We follow the hyper-parameter setup described in DUQ \cite{van2020uncertainty}. We use a standard ResNet-18 architecture, and train for 85 epochs (instead of 75), with an initial learning rate of 0.1 that is reduced by a factor of 0.2 after every 25 epochs. We simply concatenate the anchor and the residual to create a 6 channel input image, instead of a regular 3 channel image. During training, we use a simple strategy of choosing anchors by reversing the order of the batch to ensure that anchors are randomly chosen in an efficient manner (note we always use $K=1$ during training). At test time, we draw random batches from the training set as anchors for batches from the test set, resulting in a test accuracy of $93.90\% \pm 0.12$ across 3 random seeds -- obtained using $K=10$ anchors, but is generally quite stable to variations in the number of anchors. To evaluate the separation performance, we use the test set of CIFAR-10 with 10,000 images, and the test set off SVHN with 26,032 images, where both datasets utilize the same normalization strategy used during training. 
\begin{table}
\centering
\small{
\caption{\small{Single model OOD Detection on CIFAR-SVHN using $\Delta$-UQ with ResNet-18.}}
		\begin{tabular}{p{1.5in}|p{0.75in}}
		
			\hline
			
		\rowcolor[HTML]{EFEFEF} 
 \multicolumn{1}{c|}{\cellcolor[HTML]{C0C0C0}{Method}} & {AUROC $\uparrow$}  \\ \hline \hline
			  
		  ResNet-18 \shortcite{he2016deep} & $91.77 \pm 1.85$\\
		  Temp. scale & $92.20\pm 2.03$\\
		  DUQ \shortcite{van2020uncertainty} & $92.70\pm 1.30$\\
		 $\Delta$-mean & $92.31\pm 0.95$ \\
		 \name $(\Delta=X;$ trivial) & $90.49 \pm 1.1$\\
		  $\Delta$-UQ $(\Delta=X-R)$ & $\mathbf{93.39 \pm 1.18}$ \\
		  \hline
		\end{tabular}
\label{tab:cifar-svhn}
\vspace{-5pt}
}
\end{table} 

\begin{figure*}[!htb]
    \centering
    \includegraphics[width=0.95\linewidth,clip,trim=0 0 0 0]{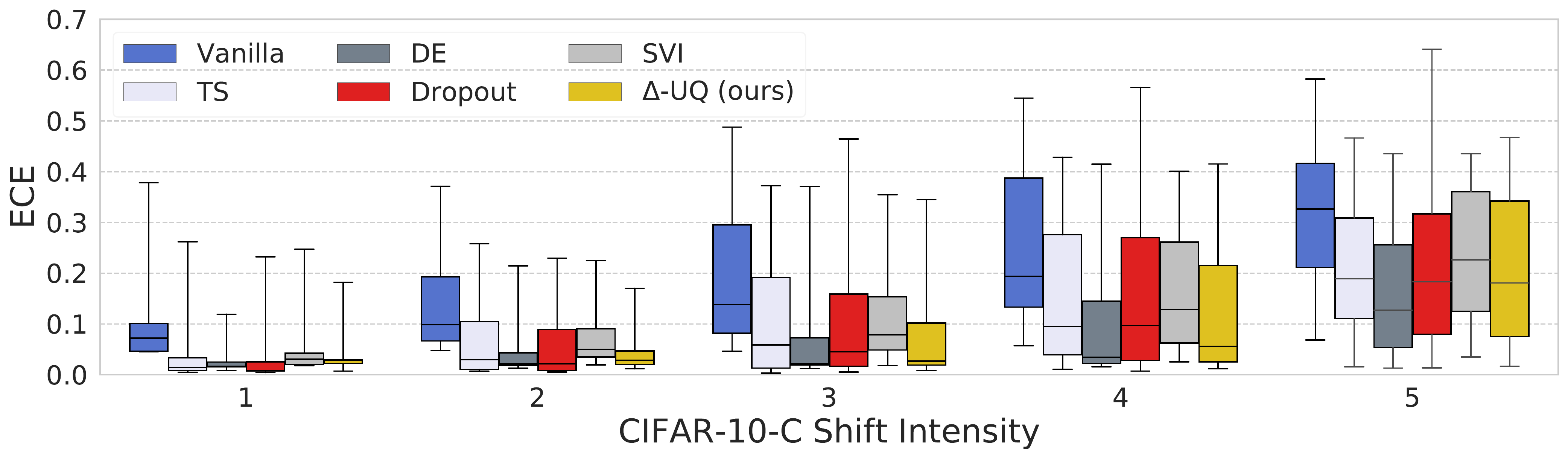}
    \caption{\small{\textbf{Expected Calibration Error (ECE $\downarrow$):} We study how \name helps in calibrating classifiers under distribution shifts between CIFAR-10 (training) and CIFAR-10-C (testing). For 16 different corruption, we report performance for the 5 intensity levels using boxplots. For each method the boxplots indicate quartiles (25, 50 and 75th percentile, and the whiskers represent the min and max values.}}
    \label{fig:calibration-error}
\end{figure*}
\paragraph{\emph{Results}} We measure performance with the area under the ROC curve in Table \ref{tab:cifar-svhn} including several baselines. To quantify uncertainty here we primarily use the commonly used predictive entropy of the softmax probabilities. We also compare the performance with other single model uncertainty estimators such as DUQ \cite{van2020uncertainty}, and temperature scaling. All the baselines are reported with models trained until 85 epochs to be consistent with ours. Although in DUQ, the models are trained only up to 75 epochs, we did not see any change of it's performance when re-training until 85 epochs.  We also report ablations of \name (a) $\Delta$-mean: predictive entropy of the mean prediction, and (b) \name ($\Delta=X-R$): our main model, that scales the logits of the mean by the uncertainty in the predictions, as $t = 0.5 - \sigma^2$, where $\sigma^2$ is the logit-wise variance of the predictions, which is the same size as the mean and scales all the logits, and (c) \name ($\Delta=X$) where we use a simpler parameterization with the same scaling as in (b). To make a fair comparison, we also scale the logits of the basic ResNet-18 model by the same constant $t = 0.5$, and report its performance as a temperature scaled (TS) version in Table \ref{tab:cifar-svhn}. Even just the mean performs comparably to temperature scaled vanilla model indicating that the $\Delta$-model training captures uncertainties accurately. Finally, we observe that the uncertainty-based scaling further improves this performance to outperform DUQ and other baselines.

\subsubsection{Calibration Under Distribution Shifts}
Next, we validate the uncertainties by studying how they can be used to calibrate classifiers under distribution shift. A model is said to be calibrated when its confidence matches its true correctness likelihood \cite{guo2017calibration}. Recently, measuring calibration error under distribution shift has been a useful metric to gauge the trustworthiness of a model's uncertainties \cite{Ovadia2019}. 

\paragraph{\emph{Experimental Results}} For OOD data, we use the corrupted version of CIFAR-10 \cite{hendrycks2018benchmarking} that contains 16 different corruptions ranging from different kinds of noise, blurs, weather changes etc. at 5 different intensities. We expect a well calibrated method to have consistently low calibration error across all five intensities. Following common evaluation criterion, we report calibration metrics split by shift intensity (1--5). To measure calibration error using commonly used metrics such as expected calibration error (ECE) \cite{naeini2015obtaining} that directly measures the difference between model accuracy and its confidence. We report other metrics like negative log likelihood (NLL), and Brier Score \cite{brier1950verification} in the supplement in addition to more details about the dataset and training settings. 

Following \cite{Ovadia2019}, we use a standard ResNet-20 \cite{he2016deep}, except the input is changed to accept 6 channels (stacking anchors and the residual) instead of the standard 3 and is trained up to 200 epochs, using a simple [0,1] normalization. We use the same normalization scheme across train/test and OOD datasets. We use a standard training scheme for ResNet-20, using an initial learning rate of 0.1, that is reduced by a factor of 0.2 after 60, 120, 160 epochs. We keep the model selection strategy simple, by using the training weights at the end of 200 epochs for our experiments. For reference, the $\Delta-$model achieves a clean test accuracy of $92.09\%\pm 0.155$ across 5 random seeds. We report the ECE performance averaged across model predictions obtained with 5 different seeds for all methods.

\paragraph{\emph{Results}} In Figure \ref{fig:calibration-error} we show boxplots for different methods including MC-Dropout~\cite{gal2016dropout}, Deep Ensembles~\cite{lakshminarayanan2016simple, Ovadia2019}, Temperature Scaling~\cite{guo2017calibration}, SVI~\cite{graves2011practical}. The boxplots indicate quartiles (25, 50 and 75th percentile, with median as the line inside) summarizing the calibration error across 16 different corruptions, and the error bars indicate the min and max values. We observe that a simple post-hoc calibration (same as before, $0.5-\sigma^2$) using \name help in calibrating classifiers across distribution shifts competitively compared to several of the baseline methods.